\documentclass[9pt,conference]{IEEEtran} 
\IEEEoverridecommandlockouts
\usepackage{cite}
\usepackage{amsmath,amssymb,amsfonts}
\usepackage{algorithmic}
\usepackage{graphicx}
\usepackage{bm}
\usepackage{multirow}
\usepackage{hyperref}
\usepackage{array}
\usepackage{cite}
\usepackage{textcomp}
\usepackage{xcolor}
\def\BibTeX{{\rm B\kern-.05em{\sc i\kern-.025em b}\kern-.08em
    T\kern-.1667em\lower.7ex\hbox{E}\kern-.125emX}}
\begin{document}

\title{Selective Attention Merging for low resource tasks: A case study of Child ASR
\thanks{This work was supported in part by the NSF.}
}

\author{
\IEEEauthorblockN{
Natarajan Balaji Shankar, 
Zilai Wang, 
Eray Eren,
Abeer Alwan}

\IEEEauthorblockA{
\textit{Department of Electrical and Computer Engineering} \\
\textit{University of California Los Angeles}\\
Los Angeles, USA \\
\textit{\{balaji1312, zilaiwang2001, erayeren\}@ucla.edu, alwan@ee.ucla.edu}
}
}
\maketitle

\begin{abstract}
While Speech Foundation Models (SFMs) excel in various speech tasks, their performance for low-resource tasks such as child Automatic Speech Recognition (ASR) is hampered by limited pretraining data. To address this, we explore different model merging techniques to leverage knowledge from models trained on larger, more diverse speech corpora. This paper also introduces Selective Attention (SA) Merge, a novel method that selectively merges task vectors from attention matrices to enhance SFM performance on low-resource tasks. Experiments on the MyST database show significant reductions in relative word error rate of up to 14\%, outperforming existing model merging and data augmentation techniques. By combining data augmentation techniques with SA Merge, we achieve a new state-of-the-art WER of 8.69 on the MyST database for the Whisper-small model, highlighting the potential of SA Merge for improving low-resource ASR.
\end{abstract}


\section{Introduction}
\label{sec:introduction}

Recently, Speech Foundation Models (SFMs) have increasingly come to dominate the landscape of Automatic Speech Recognition (ASR)\cite{chen2022wavlm, hsu2021hubert, baevski2020wav2vec,baevski2022data2vec, chung2021w2v,Rad23whisper, babu2022xls, peng2024owsm, puvvada2024less} due to their impressive performance in a range of different speech datasets, and considerable zero-shot ability. The success of SFMs can be attributed to several factors, including large-scale pretraining on diverse datasets and learning objectives that leverage unlabeled (self supervised) or weakly supervised data. However, the performance of these models in child speech related tasks still lags behind that seen in general adult speech \cite{fan2024benchmarking}. This performance gap can be primarily attributed to the significant acoustic and linguistic differences between child and adult speech, such as higher pitch, greater variability in pronunciation, and the use of simpler vocabulary and sentence structures \cite{lee1999acoustics}.
\\
\\
Several strategies have been proposed to tackle this domain mismatch. Signal processing based acoustic techniques to address the data scarcity issue involve using data augmentation techniques \cite{patel2011prosodic, ko2015audio, jaitly2013vocal, park2019specaugment} to artificially increase the variety of data seen during fine-tuning of the model. These techniques involve applying transformations to the original speech signals, such as pitch shifting, time stretching, or masking parts of the spectrogram, to create new training examples that can help the model generalize better to unseen child speech. Another way to increase the diversity of training data seen is to use synthetic data created using voice conversion or TTS on in-domain text to increase the robustness of models to child speech \cite{shahnawazuddin20_interspeech, zhao23c_interspeech, rolland2024improved}. While these methods do improve the performance of SFMs in the new domain, these often involve creating artificial data, before retraining the model on each new dataset encountered.
\\
\\
Model merging \cite{wortsman2022model, matena2022merging, yu2024language, ilharcoediting, jindataless, yadav2024ties} has recently emerged as a compelling alternative to training models with new augmented sets. By leveraging the knowledge learned by a model on more comprehensive corpora, we can improve the performance of the same model on a more limited dataset. This approach avoids the need for extensive data collection and retraining, which can be particularly beneficial in scenarios like low-resource child ASR. In addition, model merging techniques have been shown to increase performance on in-domain tasks by increasing the generalizability of the model \cite{wortsman2022model}. This suggests that model merging can both help models adapt to new domains and improve performance on existing domains without requiring explicit fine-tuning, making it a promising approach for addressing the domain mismatch in child ASR.
\\
\\
This paper presents an investigation into the application of model merging in low-resource child ASR, along with introducing a novel Selective Attention (SA) Merge technique.
Our contributions can be summarized as follows:
\begin{itemize}
	\item
	Exploring the viability of existing model merging techniques for low-resource child ASR across different SFMs
	\item 
	Introduction of a new SA Merge technique for low-resource domain adaptation of SFMs
	\item 
	Adaptation of existing data augmentation techniques to different child speech datasets through task vector transfer
\end{itemize}
The remainder of this paper is organized as follows. Section \ref{sec:methods} introduces the proposed SA Merge method, and serves as an introduction to other model merging methods tested. Experimental setups and datasets are described in Section \ref{sec:experiments}. Results are discussed in Section \ref{sec:results}, and we conclude the paper in Section \ref{sec:conclusion}.\footnote{Our code, models, and data splits are available at \url{https://github.com/balaji1312/sa\_merging}}

\section{Methods}
\label{sec:methods}

\subsection{Model Merging}
Model merging has emerged as a promising research area, aiming to combine multiple domain specific models into a single model. However, its effectiveness for child ASR remains unexplored. We present the first evaluation of these methods for merging models fine-tuned for child ASR, investigating the following techniques:

\begin{itemize}
	\item 
	\textbf{Linear Interpolation (Lerp)}\cite{wortsman2022model}: Lerp creates a merged model by computing a weighted average of the parameters from individual models.
	
	\item
	\textbf{Spherical Linear Interpolation (Slerp)}: Similar to Lerp, Slerp performs interpolation in a spherical space \cite{shoemake1985animating}, often resulting in smoother transitions between model parameters.
	
	\item
	\textbf{Task Arithmetic (TA)}\cite{ilharcoediting}: TA involves computing task-specific vectors by taking the difference between task-specific models and a pretrained base model. The vectors are combined using predefined scaling factors to adjust the relative importance of different models during the merging process.
	
	
	\item
	\textbf{Regression Mean (RegMean)}\cite{jindataless}: RegMean merging formulates model merging as an optimization problem, minimizing prediction differences between the merged model and the individual models through linear regression.
	
	\item
	\textbf{TIES Merging (TIES)}\cite{yadav2024ties}: TIES Merging tackles conflicts in model merging by trimming low-magnitude parameters, resolving sign disagreements, and merging parameters with consistent signs.
	
	\item 
	\textbf{DARE Merging (DARE)}\cite{yu2024language}: DARE can be used in combination with other merging methods, as it involves dropping a random percent of parameter differences, and rescaling the remaining weights before merging.
	
\end{itemize}

\subsection{Selective Attention Merge}
Previous analyses have shown the importance of attention maps in different layers for speech-based learning, especially for speech in low-resource \cite{shankar2024soa} and noisy \cite{lee2022regularizing} environments. Building on these insights, we propose a novel approach called Selective Attention (SA) Merge, which focuses on merging the task vectors of attention matrices while preserving the weights of other layers.

Drawing inspiration from \cite{sundar2024multimodal} on cross-modal merging techniques and \cite{stolcke1994} on model merging within Hidden Markov Models (HMMs) for low-resource domains, we propose a novel approach to transfer knowledge from a comprehensive source domain to the low-resource target domain. While \cite{stolcke1994} focused on merging speech models using HMMs, our approach centers around the merging of attention matrices. SA Merge combines the task vectors of attention matrices from two models, $\mathcal{M}_{1}$ (fine-tuned on child speech) and $\mathcal{M}_{2}$ (fine-tuned on diverse adult speech), as follows:
\begin{equation}
	{}_{\mathcal{M}_{SA}}\tau_{i}^{Q,K,V} = \lambda_{i} \cdot {}_{\mathcal{M}_{1}}\tau_{i}^{Q,K,V} + (1 - \lambda_{i}) \cdot {}_{\mathcal{M}_{2}}\tau_{i}^{Q,K,V}
\end{equation}
where
${}_{\mathcal{M}_{SA}}\tau_{i}^{Q,K,V}$ represents the merged task vectors for the query, key, and value matrices in the $i$-th attention layer of the new model $\mathcal{M}_{SA}$,    ${}_{\mathcal{M}_{1}}\tau_{i}^{Q,K,V}$ and ${}_{\mathcal{M}_{2}}\tau_{i}^{Q,K,V}$ are the corresponding task vectors from the child speech and adult speech models respectively, and $\lambda_{i}$ is the mixing ratio for the $i$-th layer, controlling the contribution of each model.

The mixing ratio $\lambda_{i}$ is further defined as:
\begin{equation}
	\lambda_{i} = \lambda^{\alpha_i}
\end{equation}
where
$\lambda$ is a base mixing factor and $\alpha_{i}$ is an exponent that controls the rate of change of the mixing ratio across layers.

By weighting the mixing ratio in an exponential manner, we aim to give higher importance to the lower layers from the child speech model $\mathcal{M}_{1}$. This is motivated by the assumption that lower layers capture more acoustic and phonetic features, which are crucial for distinguishing child speech from adult speech. At higher layers, the influence of the adult speech model $\mathcal{M}_{2}$ gradually increases, allowing the merged model to benefit from the broader linguistic knowledge captured in the adult speech data. Unlike \cite{sundar2024multimodal}, where model merging is explored between attention matrices from models trained on different modalities, we apply task vector based merging, taking into account the amount of target domain data learned by each model through an exponential weighting.
For non-attention layers, we retain the weights from the target domain fine-tuned model, i.e., the child speech model $\mathcal{M}_{1}$. This ensures that the merged model retains the specialized knowledge acquired from the child speech data, while benefiting from the capabilities of the adult speech model $\mathcal{M}_{2}$. 

\subsection{Task Vector Transfer}\label{tv}
Recent advancements in applying task vectors to speech have demonstrated remarkable success in transferring learned augmentations across datasets \cite{su2024taskarithmeticmitigatesynthetictoreal}. This process typically involves calculating the difference between task vectors derived from two models: $\mathcal{M}$ trained on the target dataset $\mathcal{D}$, and $\mathcal{M}^{'}$ trained on an augmented version of the dataset $\mathcal{D^{'}}$.
\begin{equation}
	\tau_i = \theta_{i,1} - \theta_{i,2} \; \; \forall \; \theta_{i,1} \in \mathcal{M}, \;\;\theta_{i,2} \in \mathcal{M}^{'}
\end{equation}
where
$\tau_i$ is the calculated task vector $i$-th attention layer of the new model, and $\theta_{i,1}$ and $\theta_{i,2}$ are the corresponding parameters from $\mathcal{M}$ trained on  $\mathcal{D}$, and $\mathcal{M}^{'}$ trained on $\mathcal{D^{'}}$.

We extend this concept by investigating the transferability of these learned task vectors to models fine-tuned on different child speech datasets. Our goal is to assess whether the performance gains observed on the source dataset can be replicated in a new setting without any additional data augmentation. Furthermore, recognizing that the vector difference operation may not entirely eliminate all characteristics learned from the source dataset $\mathcal{D}$, we also evaluate the performance of these task vectors in a zero-shot setting to provide insights into the extent to which the learned task vectors capture generalizable knowledge about child speech patterns.

\section{Experiments}
\label{sec:experiments}

\subsection{Experimental Setup}
To assess the effectiveness of various merging techniques, including our proposed SA Merge, we conducted experiments across a diverse range of supervised and self-supervised speech foundation models (SFMs). Specifically, we evaluated several models from the Whisper family \cite{Rad23whisper} of varying sizes, using the Hugging Face Transformers \cite{wolf2020transformers} library for fine-tuning.
Among SSL models, we evaluated the base versions of Wav2Vec2.0 \cite{baevski2020wav2vec}, HuBERT \cite{hsu2021hubert} and WavLM \cite{chen2022wavlm}. All SSL models were trained with an identical character level CTC loss based on their implementation in the fairseq toolkit \cite{ott2019fairseq}. 
To offer a comparison of model merging techniques, we evaluate Lerp, Slerp, TA, RegMean, TIES, and DARE, in addition to our proposed SA Merge. The hyperparameters $\alpha$ and $\lambda$ for SA Merge were tuned within the ranges [0.7, 0.9] and [0.1, 0.3] respectively. Our implementation of SA Merge, as well as the evaluation of the different models, was facilitated by the Mergekit library \cite{goddard2024arcee}. All models listed were fine-tuned using 2 Nvidia A6000 GPUs.

\subsection{Datasets}

For our merging experiments, we fine-tune two models: $\mathcal{M}_{1}$ on low-resource child speech and $\mathcal{M}_{2}$ on mainstream speech. To obtain $\mathcal{M}_{2}$, we train a given pretrained base model on the train-100-hour subset of the LibriSpeech (LS) corpus \cite{PanayotovCPK15}. For $\mathcal{M}_{1}$, we utilize the following datasets:
\subsubsection{My Science Tutor}

The My Science Tutor (MyST) corpus \cite{ward2011my} comprises approximately 240 hours of transcribed conversational children's speech, spanning grades 3 to 5, collected during virtual tutoring sessions on subjects including physics, geography, biology, and other science topics. Similar to~\cite{attia2023kid}, we identify and filter low quality audio samples by removing utterances with WER larger than 50\% (after passing through Whisper-large-v2) or with less than 3 words are removed. Utterances longer than 30s are also removed in both the training and test sets, resulting in filtered data splits as follows: train (133h), dev (21h), and test (25h). To verify the efficacy of the proposed method under more constrained settings, we also separately prepare 1-hour, 5-hour, and 10-hour subsets of the MyST train corpus.

\subsubsection{CMU Kids}

To demonstrate the transferability of task vectors across different children's speech datasets, we evaluated performance on the CMU Kids Corpus \cite{eskenazi1997cmu}. This corpus consists of 5180 utterances of read speech from 76 speakers, totaling 9 hours of child speech data. The utterances are randomly partitioned into train (70\%), development (15\%), and test (15\%) sets, ensuring no speaker overlap between the sets.

\subsubsection{Data Augmentations}

To provide a contrast to the use of model merging, we compare several widely employed data augmentation methods on the MyST corpus. These include pitch perturbation (PP)~\cite{patel2011prosodic}, speed perturbation (SP)~\cite{ko2015audio}, vocal tract length perturbation (VTLP)~\cite{jaitly2013vocal}, and SpecAugment (SpecAug)~\cite{park2019specaugment}. In addition to the above methods, we also generate synthetic TTS data on the MyST corpus.

\begin{itemize}
	\item \textbf{Pitch perturbation (PP)} The fundamental frequency of each utterance is randomly shifted up or down by 1 to 12 semitones, creating two additional copies.
	\item \textbf{Speed perturbation (SP)} The speed of each utterance is modified, creating two copies with perturbation rates of 0.9 and 1.1.
	\item \textbf{Vocal tract length perturbation (VTLP)} This technique applies frequency warping to the speech signal, creating two copies with perturbation rates of 0.9 and 1.1.
	\item \textbf{SpecAugment (SpecAug)} Random masking of consecutive frequency bands and time frames is applied while training.
	\item \textbf{Synthetic Data}. Synthetic data is generated using StyleTTS 2 \cite{li2024styletts} with cross-utterance text, doubling the training data. To avoid contamination, only speakers in the respective training subset are used for synthetic data generation.
\end{itemize}

\section{Results}
\label{sec:results}

\subsection{Can Model Merging facilitate Knowledge Transfer in Low Resource Child ASR?}

\subsubsection{Evaluation of Model Merging techniques on Supervised SFMs}
We first examine the effectiveness of different merging techniques to facilitate knowledge transfer using the Whisper-small model. For this purpose, we combine a model fine-tuned on the 100-hour subset of the LibriSpeech (LS) corpus \cite{PanayotovCPK15} with models fine-tuned on the 1-hour, 5-hour, 10-hour, and the full train subset of the MyST corpus \cite{ward2011my}. We conduct a hyperparameter search for each model merging method and report the best results in Table \ref{tab:tab1}. Our results demonstrate that the proposed SA Merge technique provides the most reliable improvements in reducing Word Error Rate (WER) across various data subsets. All subsequent results in this section show a statistically significant ($p < 0.05$) improvement for SA Merge compared to the baseline zero-shot performance of the models.

\begin{table}[!htbp]\centering
	\caption{WER results on MyST test set from merging Whisper-small models trained on MyST and LS train-100. Zero shot denotes a model without fine-tuning. ft denotes a model fine-tuned on the respective MyST subset without any merging. Bold face numbers indicate best results.}\label{tab:tab1}
	\resizebox{\columnwidth}{!}{%
		\begin{tabular}{|c|c|c|c|c|}
			\hline
			Merging&\multicolumn{4}{c|}{MyST test WER} \\\cline{2-5} Method &MyST 1hr &MyST 5hr  &MyST 10hr & MyST full\\ \hline\hline
			Zero Shot & \multicolumn{4}{c|}{13.44}\\ \hline
			
			ft & 10.64& 10.05& 9.94& 9.34  \\
			\hline\hline
			
			Lerp & 10.51 & 9.94& 10.05& 8.86 \\
			\hline
			Slerp & 10.51& 9.94& 10.05&  8.88\\
			\hline
			TA & 10.60& 10.03& 10.11&  9.10 \\
			\hline
			DARE + TA & 10.43& 10.07& 10.16&  9.16\\
			\hline
			TIES & 10.70& 9.97& 10.00&  8.92  \\
			\hline
			SA Merge (Ours) & \textbf{10.40}& \textbf{9.85} & \textbf{9.80}& \textbf{8.85}\\
			\hline
		\end{tabular}%
	}
\end{table}
			
			

Next, we examine the impact of model merging when the size of the supervised SFM is varied. We train supervised SFMs of different sizes from the Whisper family on the 1-hour, 5-hour, 10-hour, and the full train subset of the MyST corpus. We then analyze the effect of merging these models with a model fine-tuned on the LS train-100 subset in Table \ref{tab:tab3}. These tests are performed using the best-performing method from Table \ref{tab:tab1} (SA Merge), but we note that similar trends are observed with other common merging methods. Generally, we find that as the available training data decreases or the model size shrinks, model merging enhances model robustness. Notably, SA Merge achieves a 14\% reduction in relative WER on a Whisper-base model fine-tuned on the MyST train 1-hour subset, highlighting its efficacy in extremely low-resource scenarios. 



\begin{table}[!htbp]\centering
	\caption{WER results on MyST test set from merging supervised SFMs trained on MyST subsets and LS train-100 using SA Merge. Bold face numbers indicate best results.}\label{tab:tab3}
	\resizebox{\columnwidth}{!}{%
		\begin{tabular}{|c|c|c|c|c|c|}
			\hline
			\multirow{2}{*} {Model}&Merging&\multicolumn{4}{c|}{MyST test WER} \\\cline{3-6} &Method &MyST 1hr &MyST 5hr  &MyST 10hr & MyST full\\ \hline\hline
			Whisper& ft & 16.12& 15.23& 14.36& 11.63\\\cline{2-6}
			tiny & SA Merge & 15.26& 14.38& 14.00& 11.52  \\\hline
			Whisper & ft & 14.95& 12.62& 12.15& 10.33\\ \cline{2-6}
			base & SA Merge & 12.84& 12.30& 11.55& 9.87\\\hline
			Whisper & ft & 9.92 & 9.42 & 9.19 & 8.86 \\ \cline{2-6}
			medium & SA Merge & 9.82 & 9.28 & 9.10 & \textbf{8.63}  \\\hline
			Whisper  & ft & 11.31 & 9.54 & 9.42 & 9.12 \\ \cline{2-6}
			large v3& SA Merge & \textbf{9.41} & \textbf{9.14} & \textbf{9.03} & 8.74  \\\hline
		\end{tabular}%
	}
\end{table}

\subsubsection{Comparison with other Data Augmentation Techniques}
Table \ref{tab:tab6} presents a comparison of common data augmentation techniques applied to various MyST subsets. As data augmentation necessitates retraining the model with the newly augmented data, whereas SA Merge operates on existing models, we also investigate the potential benefits of combining these two approaches. Our results indicate that across all data augmentation methods, the application of SA Merge consistently improves model performance on the MyST test set. Notably, utilizing SA Merge in conjunction with SpecAug yields a WER of 8.69, establishing a new state-of-the-art performance on the MyST dataset for the Whisper-small model.

\begin{table}[!htbp]\centering
	\caption{WER results on MyST test set from merging Whisper-small models trained on augmented MyST data and LS train-100 using SA Merge. PP, SP, VTLP, SpecAug denote data augmentations. TTS and Real indicate pure synthetic TTS data and original MyST data. Bold face numbers indicate best results.}\label{tab:tab6}
	\resizebox{\columnwidth}{!}{%
		\begin{tabular}{|c|c|c|c|c|}
			\hline
			Augmentation/&\multicolumn{4}{c|}{MyST test WER} \\\cline{2-5} Merging Method &MyST 1hr &MyST 5hr  &MyST 10hr & MyST full\\ \hline\hline
			PP & 10.21 & 9.99 & 9.62 & 8.84\\\hline
			PP + SA Merge & 9.65 & 9.29 & 9.32 & 8.80\\\hline\hline
			SP & 10.04 & 10.72 & 10.38 & 8.89\\\hline
			SP + SA Merge & 9.53 & 9.52 & 9.14 & 9.01\\\hline\hline
			VTLP & 9.91 & 9.64 & 9.18 & 8.95\\\hline
			VTLP + SA Merge & 9.63 & 9.24 & 9.04 & 8.75\\\hline\hline
			SpecAug & 9.77 & 9.35 & 9.25 & 9.03\\\hline
			SpecAug + SA Merge & 9.54 & 9.21 & 9.05 & \textbf{8.69}\\\hline\hline
			TTS & 13.23& 12.04& 12.19& 12.61\\\hline
			TTS + Real & 9.13& 9.25& 9.30& 8.89\\\hline
			TTS + Real + SA Merge & \textbf{8.84}& \textbf{8.85}& \textbf{8.84}& 8.74\\\hline
		\end{tabular}%
	}
\end{table}

\subsubsection{Comparison of Model Merging Techniques for Self Supervised SFMs}
For completeness, we also examine the effectiveness of merging methods on different Self Supervised SFMs in Table \ref{tab:tab5}. In line with the findings of \cite{fan2024benchmarking}, we note a general trend of self-supervised SFMs exhibiting higher WER on child ASR tasks compared to their supervised counterparts. We also observe that task vector-based methods (including SA Merge) underperform direct parameter merging techniques like Lerp, although they still outperform fine-tuned models without any merging. We hypothesize that this discrepancy may be attributed to the significant task shift from the SSL objective to the CTC objective during the fine-tuning of Self Supervised SFMs. However, a more in-depth exploration of this phenomenon is left for future work.

\begin{table}[!htbp]\centering
	\caption{WER results on MyST test set from merging self supervised SFMs trained on MyST subsets and LS train-100 using Lerp and SA Merge. Bold face numbers indicate best results.}\label{tab:tab5}
	\resizebox{\columnwidth}{!}{%
		\begin{tabular}{|c|c|c|c|c|c|}
			\hline
			\multirow{2}{*}{Model} &Merging&\multicolumn{4}{c|}{MyST test WER} \\\cline{3-6}&Method &MyST 1hr &MyST 5hr  &MyST 10hr & MyST full\\ \hline\hline
			\multirow{3}{*}{Wav2Vec2.0}& ft & 31.95 & 21.05& 18.43 & 13.18\\\cline{2-6}
			& Lerp & 30.40 & 19.87 & 17.77 & 12.83 \\\cline{2-6}
			& SA Merge & 31.33 & 20.57 & 19.90 & 13.14 \\\hline
			\multirow{3}{*}{HuBERT}& ft & 34.78 & 21.81 & 19.29 & 13.30\\ \cline{2-6}
			& Lerp & 32.09 & 20.60 & 18.07 & 13.01\\ \cline{2-6}
			& SA Merge & 32.47 & 20.65 & 18.09 & 12.78  \\\hline
			\multirow{3}{*}{WavLM}& ft & 27.37 & 18.03 & 15.86 & 12.38\\ \cline{2-6}
			& Lerp & \textbf{25.84} & \textbf{16.56} & \textbf{14.93} & \textbf{12.08}  \\\cline{2-6}
			& SA Merge & 26.52 & 17.41 & 15.28 & 12.23  \\\hline
		\end{tabular}%
	}
\end{table}



\subsection{Can we isolate Task Vectors from different data augmentation techniques?}
\subsubsection{Transferability of Data Augmentation Task Vectors}

In addition to our investigations into the direct merging of models, we also intend to compute the task vectors for models that have been fine-tuned using various data augmentation techniques, as discussed in Section \ref{tv}. Subsequently, we investigate whether these computed task vectors can be effectively transferred to other datasets to enhance their performance. Different from \cite{su2024taskarithmeticmitigatesynthetictoreal}, we do not compute an overall task vector based on performance across all tasks; instead, we focus on deriving task vectors specific to the particular dataset.

We begin by fine-tuning Whisper-small models on the entire MyST train corpus, incorporating common augmentations used in child ASR (PP, SP, SA, VTLP), as well as synthetic data. Subsequently, we compute the difference in task vectors between these models and transfer these differences to two Whisper-small models: one in a zero-shot setting and another fine-tuned on the CMU Kids corpus. Our results, presented in Table \ref{tab:tab8}, reveal that task vector transfer leads to a relative WER reduction of 18\% in the zero-shot setting and 11\% in the fine-tuned model, eliminating the need for retraining on augmented data for the new dataset.

\begin{table}[!htbp]\centering
	\caption{WER on CMU Kids test set from transferring task vectors (tv) of augmentations (PP, VTLP, SP, SpecAug, TTS) from Whisper-small fine-tuned on MyST to both zero-shot (no fine-tuning) and CMU Kids fine-tuned models. Bold face numbers indicate best results.}\label{tab:tab8}
	\resizebox{0.6\columnwidth}{!}{%
		\begin{tabular}{|c|c|c|}
			\hline
			Merging & \multicolumn{2}{c|}{CMU Kids test WER}\\\cline{2-3} Method  &zero-shot & fine-tuned \\
			\hline\hline
			None & 11.36 & 2.01	 \\
			\hline\hline
			
			VTLP tv & 9.52 & 1.91 \\
			\hline
			PP tv&9.66 & \textbf{1.79}  \\
			\hline
			SpecAug tv & 9.54 & 1.83\\
			\hline
			SP tv & 9.48 & 1.86\\
			\hline
			TTS tv & \textbf{9.29} & 1.85\\
			\hline
		\end{tabular}%
	}
	
\end{table}

\subsubsection{Alignment of Data Augmentation Task Vectors}

Building on the notion that task vectors encapsulate both task-specific information (which boost performance) and inherent robustness, we analyze the pairwise cosine similarities between the computed task vectors, following a similar approach to \cite{su2024taskarithmeticmitigatesynthetictoreal}.

\begin{figure}[htp]
	\centering{{\includegraphics[width=0.47\textwidth]{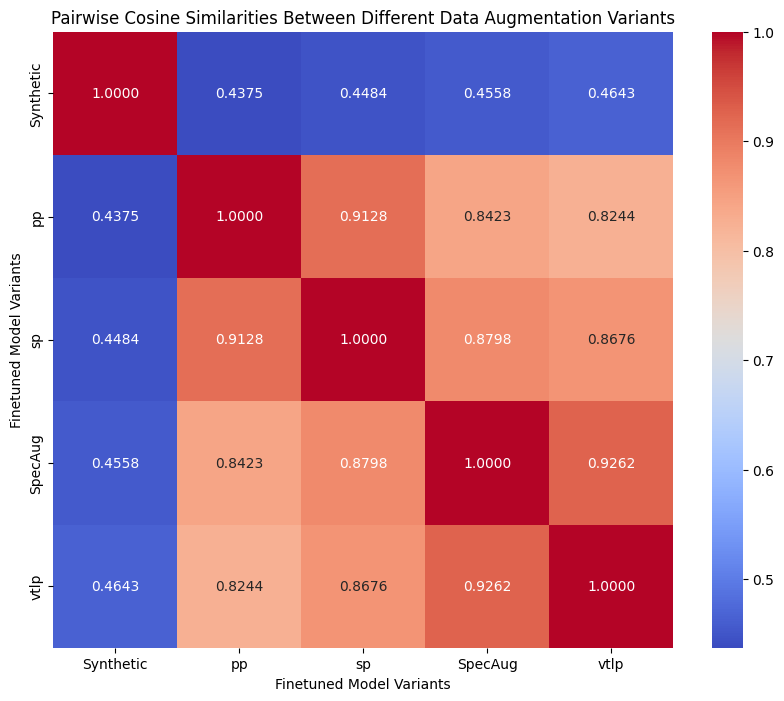} }}%
	\caption{Pairwise cosine similarities between task vectors derived from Whisper-small models fine-tuned on various MyST data augmentations.}%
	\label{fig:fig1}
\end{figure}

Our analysis in Figure \ref{fig:fig1} reveals that task vectors derived from conventional signal processing-based data augmentation techniques exhibit high similarity, suggesting that the performance gains from combining these techniques might be limited, as observed by \cite{fan2024benchmarking}. In contrast, the low alignment of vectors from synthetic data indicates that this method could potentially be used in conjunction with other techniques to further enhance performance.

\section{Conclusion}
\label{sec:conclusion}

In this paper, we explored the potential of model merging to enhance low-resource child automatic speech recognition (ASR) without the need to retrain models on new datasets. We introduced Selective Attention (SA) Merge, a novel technique that selectively merges task vectors from attention matrices to improve the performance of speech foundation models (SFMs) on child speech data. Our experiments demonstrated that model merging, particularly SA Merge, leads to significant improvements in low-resource scenarios, achieving relative word error rate (WER) reductions of up to 14\%. By combining data augmentation techniques with SA Merge, we achieve a new state-of-the-art WER of 8.69 on the MyST database for the Whisper-small model. Further analysis of task vectors revealed the transferability of learned augmentations across datasets and the potential for combining multiple data augmentation techniques to further enhance child ASR systems. These findings underscore the efficacy of model merging, and SA Merge in particular, as a promising approach for addressing the challenges of child ASR in resource-constrained environments. Future research will explore the effectiveness of model merging in other low-resource domains, expanding its potential benefits beyond child ASR.

\footnotesize
\bibliographystyle{IEEEbib}
\bibliography{refs}

\begin{thebibliography}{10}

\bibitem{chen2022wavlm}
Sanyuan Chen et~al.,
\newblock ``Wavlm: Large-scale self-supervised pre-training for full stack
  speech processing,''
\newblock {\em IEEE Journal of Selected Topics in Signal Processing}, vol. 16,
  no. 6, pp. 1505--1518, 2022.

\bibitem{hsu2021hubert}
Wei-Ning Hsu et~al.,
\newblock ``Hubert: Self-supervised speech representation learning by masked
  prediction of hidden units,''
\newblock {\em IEEE/ACM Transactions on Audio, Speech, and Language
  Processing}, vol. 29, pp. 3451--3460, 2021.

\bibitem{baevski2020wav2vec}
Alexei Baevski et~al.,
\newblock ``Wav2vec 2.0: A framework for self-supervised learning of speech
  representations,''
\newblock {\em Advances in Neural Information Processing Systems}, vol. 33, pp.
  12449--12460, 2020.

\bibitem{baevski2022data2vec}
Alexei Baevski et~al.,
\newblock ``Data2vec: A general framework for self-supervised learning in
  speech, vision and language,''
\newblock in {\em International Conference on Machine Learning}. PMLR, 2022,
  pp. 1298--1312.

\bibitem{chung2021w2v}
Yu-An Chung et~al.,
\newblock ``W2v-bert: Combining contrastive learning and masked language
  modeling for self-supervised speech pre-training,''
\newblock in {\em 2021 IEEE Automatic Speech Recognition and Understanding
  Workshop (ASRU)}. IEEE, 2021, pp. 244--250.

\bibitem{Rad23whisper}
Alec Radford et~al.,
\newblock ``Robust speech recognition via large-scale weak supervision,''
\newblock in {\em International Conference on Machine Learning, {ICML} 2023}.
  2023, vol. 202, pp. 28492--28518, {PMLR}.

\bibitem{babu2022xls}
Arun Babu et~al.,
\newblock ``Xls-r: Self-supervised cross-lingual speech representation learning
  at scale,''
\newblock in {\em Proc. Interspeech 2022}, 2022.

\bibitem{peng2024owsm}
Yifan Peng et~al.,
\newblock ``Owsm v3.1: Better and faster open whisper-style speech models based
  on e-branchformer,''
\newblock {\em arXiv preprint arXiv:2401.16658}, 2024.

\bibitem{puvvada2024less}
Krishna~C Puvvada et~al.,
\newblock ``Less is more: Accurate speech recognition \& translation without
  web-scale data,''
\newblock in {\em Proc. Interspeech 2024}, 2024.

\bibitem{fan2024benchmarking}
Ruchao Fan et~al.,
\newblock ``Benchmarking children's asr with supervised and self-supervised
  speech foundation models,''
\newblock in {\em Proc. Interspeech 2024}, 2024.

\bibitem{lee1999acoustics}
Sungbok Lee et~al.,
\newblock ``Acoustics of children’s speech: Developmental changes of temporal
  and spectral parameters,''
\newblock {\em The Journal of the Acoustical Society of America}, vol. 105, no.
  3, pp. 1455--1468, 1999.

\bibitem{patel2011prosodic}
Rupal Patel et~al.,
\newblock ``Prosodic adaptations to pitch perturbation in running speech,''
\newblock {\em Journal of speech, language, and hearing research : JSLHR},
  2011.

\bibitem{ko2015audio}
Tom Ko et~al.,
\newblock ``Audio augmentation for speech recognition,''
\newblock {\em Sixteenth annual conference of the international speech
  communication association}, 2015.

\bibitem{jaitly2013vocal}
Navdeep Jaitly and Geoffrey~E Hinton,
\newblock ``Vocal tract length perturbation (vtlp) improves speech
  recognition,''
\newblock {\em Proc. ICML Workshop on Deep Learning for Audio, Speech and
  Language}, vol. 117, pp. 21, 2013.

\bibitem{park2019specaugment}
Daniel~S. Park et~al.,
\newblock ``Specaugment: {A} simple data augmentation method for automatic
  speech recognition,''
\newblock in {\em Proc. Interspeech 2019}, 2019, pp. 2613--2617.

\bibitem{shahnawazuddin20_interspeech}
S.~Shahnawazuddin et~al.,
\newblock ``{Voice Conversion Based Data Augmentation to Improve Children’s
  Speech Recognition in Limited Data Scenario},''
\newblock in {\em Proc. Interspeech 2020}, 2020, pp. 4382--4386.

\bibitem{zhao23c_interspeech}
Shuyang Zhao et~al.,
\newblock ``{Data augmentation for children ASR and child-adult speaker
  classification using voice conversion methods},''
\newblock in {\em Proc. Interspeech 2023}, 2023, pp. 4593--4597.

\bibitem{rolland2024improved}
Thomas Rolland and Alberto Abad,
\newblock ``Improved children’s automatic speech recognition combining
  adapters and synthetic data augmentation,''
\newblock in {\em ICASSP 2024-2024 IEEE International Conference on Acoustics,
  Speech and Signal Processing (ICASSP)}. IEEE, 2024, pp. 12757--12761.

\bibitem{wortsman2022model}
Mitchell Wortsman et~al.,
\newblock ``Model soups: averaging weights of multiple fine-tuned models
  improves accuracy without increasing inference time,''
\newblock in {\em International conference on machine learning}. PMLR, 2022,
  pp. 23965--23998.

\bibitem{matena2022merging}
Michael~S Matena and Colin~A Raffel,
\newblock ``Merging models with fisher-weighted averaging,''
\newblock {\em Advances in Neural Information Processing Systems}, vol. 35, pp.
  17703--17716, 2022.

\bibitem{yu2024language}
Le~Yu et~al.,
\newblock ``Language models are super mario: Absorbing abilities from
  homologous models as a free lunch,''
\newblock in {\em Forty-first International Conference on Machine Learning},
  2024.

\bibitem{ilharcoediting}
Gabriel Ilharco et~al.,
\newblock ``Editing models with task arithmetic,''
\newblock in {\em The Eleventh International Conference on Learning
  Representations}, 2023.

\bibitem{jindataless}
Xisen Jin et~al.,
\newblock ``Dataless knowledge fusion by merging weights of language models,''
\newblock in {\em The Eleventh International Conference on Learning
  Representations}, 2023.

\bibitem{yadav2024ties}
Prateek Yadav et~al.,
\newblock ``Ties-merging: Resolving interference when merging models,''
\newblock {\em Advances in Neural Information Processing Systems}, vol. 36,
  2024.

\bibitem{shoemake1985animating}
Ken Shoemake,
\newblock ``Animating rotation with quaternion curves,''
\newblock in {\em Proceedings of the 12th annual conference on Computer
  graphics and interactive techniques}, 1985, pp. 245--254.

\bibitem{shankar2024soa}
Natarajan~Balaji Shankar, Ruchao Fan, and Abeer Alwan,
\newblock ``Soa: Reducing domain mismatch in ssl pipeline by speech only
  adaptation for low resource asr,''
\newblock in {\em 2024 IEEE International Conference on Acoustics, Speech, and
  Signal Processing Workshops (ICASSPW)}, 2024, pp. 560--564.

\bibitem{lee2022regularizing}
Mun-Hak Lee et~al.,
\newblock ``Regularizing transformer-based acoustic models by penalizing
  attention weights for robust speech recognition,''
\newblock in {\em Proc. Interspeech 2022}, 2022, pp. 56--60.

\bibitem{sundar2024multimodal}
Anirudh~S Sundar et~al.,
\newblock ``Multimodal attention merging for improved speech recognition and
  audio event classification,''
\newblock in {\em 2024 IEEE International Conference on Acoustics, Speech, and
  Signal Processing Workshops (ICASSPW)}. IEEE, 2024, pp. 655--659.

\bibitem{stolcke1994}
A.~Stolcke and S.M. Omohundro,
\newblock ``Best-first model merging for hidden markov model induction,''
\newblock Tech. {R}ep., Tech. Rep. TR–94–003, ICSI, University of
  California, Berkeley., 1994.

\bibitem{su2024taskarithmeticmitigatesynthetictoreal}
Hsuan Su et~al.,
\newblock ``Task arithmetic can mitigate synthetic-to-real gap in automatic
  speech recognition,''
\newblock {\em arXiv preprint arXiv:2406.02925}, 2024.

\bibitem{wolf2020transformers}
Thomas Wolf et~al.,
\newblock ``Transformers: State-of-the-art natural language processing,''
\newblock in {\em Proceedings of the 2020 conference on empirical methods in
  natural language processing: system demonstrations}, 2020, pp. 38--45.

\bibitem{ott2019fairseq}
Myle Ott et~al.,
\newblock ``fairseq: A fast, extensible toolkit for sequence modeling,''
\newblock {\em Proceedings of the 2019 Conference of the North American Chapter
  of the Association for Computational Linguistics}, pp. 48--53, 2019.

\bibitem{goddard2024arcee}
Charles Goddard et~al.,
\newblock ``Arcee's mergekit: A toolkit for merging large language models,''
\newblock {\em arXiv preprint arXiv:2403.13257}, 2024.

\bibitem{PanayotovCPK15}
Vassil Panayotov et~al.,
\newblock ``Librispeech: An {ASR} corpus based on public domain audio books,''
\newblock {\em ICASSP 2015-2015 IEEE International Conference on Acoustics,
  Speech and Signal Processing (ICASSP)}, pp. 5206--5210, 2015.

\bibitem{ward2011my}
Wayne Ward et~al.,
\newblock ``My science tutor: A conversational multimedia virtual tutor for
  elementary school science,''
\newblock {\em ACM Transactions on Speech and Language Processing (TSLP)}, vol.
  7, no. 4, pp. 1--29, 2011.

\bibitem{attia2023kid}
Ahmed~Adel Attia et~al.,
\newblock ``Kid-whisper: Towards bridging the performance gap in automatic
  speech recognition for children vs. adults,''
\newblock {\em arXiv preprint arXiv:2309.07927}, 2023.

\bibitem{eskenazi1997cmu}
Maxine Eskenazi et~al.,
\newblock ``The cmu kids corpus,''
\newblock {\em Linguistic Data Consortium}, vol. 11, 1997.

\bibitem{li2024styletts}
Yinghao~Aaron Li et~al.,
\newblock ``Styletts 2: Towards human-level text-to-speech through style
  diffusion and adversarial training with large speech language models,''
\newblock {\em Advances in Neural Information Processing Systems}, vol. 36,
  2024.

\end{thebibliography}

\end{document}